\documentclass[letterpaper, 10 pt, conference]{ieeeconf}

\IEEEoverridecommandlockouts
\newif\ifcamera

\cameratrue       

\usepackage{graphics}
\usepackage{epsfig}
\usepackage{mathptmx}
\usepackage{times}
\usepackage{amsmath}
\usepackage{amssymb}
\usepackage{booktabs}
\usepackage{multirow}

\usepackage{capt-of} 

\title{\LARGE \bf
VGM-VS: Rethinking Visual Geometry Model for High-Precision Visual Servoing}

\ifcamera

\author{
Yimin Pan$^{1}$\quad Sen Wang$^{1,2}$ \quad You Zhou$^{1}$ \quad Jianfeng Gao$^{1}$ \quad Pengbo Sun$^{1}$ \quad \\  Ahmed M.~Naguib$^{1}$ \quad  Zoltan-Csaba Marton$^{1}$%
\thanks{ $^1$ All authors are with Agile Robots SE, Munich, Germany.}%
\thanks{ $^2$ Technical University of Munich}
}
\else

\author{Anonymous Authors}

\fi

\IEEEaftertitletext{%
	\vspace{-0.04in}%
	\noindent
	\begin{minipage}{\textwidth}
		\centering
		\includegraphics[width=\textwidth]{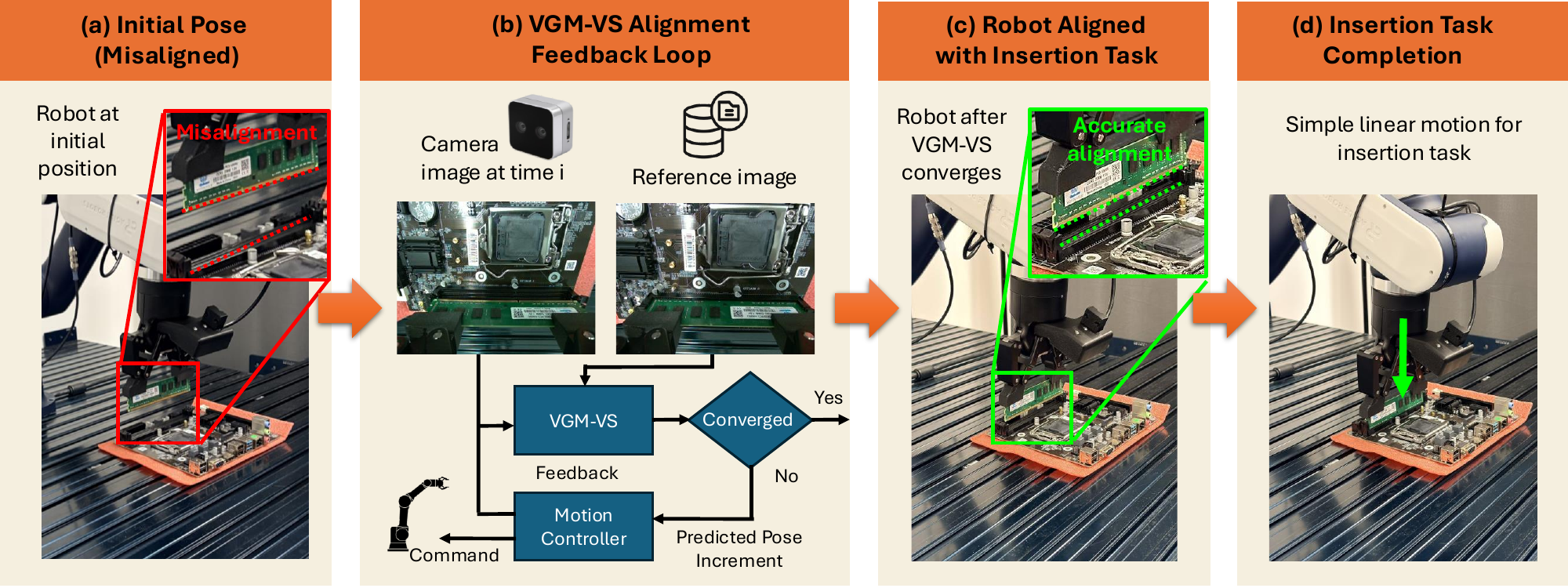}
        \vspace{-7mm}
		\captionof{figure}{Overview of VGM-VS, shown on the RAM insertion task.
        (a)~The robot starts from an arbitrary viewpoint, misaligned with the target.
        (b)~VGM-VS predicts the relative pose between the current and the reference image,
        which a motion controller applies in a closed loop until convergence.
        (c)~The loop ends at a high-precision alignment, (d)~from which a short linear
        motion completes the task.}
		\label{fig:teaser}
	\end{minipage}
	\vspace{3mm}%
}

\begin{document}
\bstctlcite{IEEEexample:BSTcontrol}

\maketitle
\thispagestyle{empty}
\pagestyle{empty}


\begin{abstract}

    We present VGM-VS, a visual servoing method built on a pretrained feed-forward
    visual geometry model. Given the current view and a reference image captured at
    the target configuration, we estimate the relative camera pose with a visual geometry model and apply it iteratively as the pose increment of a closed-loop pose-based visual servoing (PBVS) scheme. The geometry-aware representation acquired from large-scale pretraining keeps this estimate reliable when the target is occluded, weakly textured, or covers only a small part of the image. However, the scale ambiguity inherent to these models leaves the predicted translation defined up to an unknown scale, while the pose increment must be metric for robot control. We close this gap with a scene-specific metric adaptation: the robot autonomously records image--pose pairs along a predefined motion starting from the target pose, and we fine-tune the camera head on these data, jointly learning the hand--eye transform and thus removing the need for a dedicated calibration process. We evaluate our method on three real-world assembly tasks with demanding tolerances: USB-C cable picking, cable insertion, and RAM insertion. Running in real time at 30Hz, VGM-VS converges to submillimeter terminal accuracy on the cable tasks, and reaches success rates of 90--100\% when the target is moved during servoing. It converges in all trials under initial displacements of up to 30cm from the reference pose and with 50\% of the target object occluded, outperforming the compared visual servoing baselines.

\end{abstract}

\section{Introduction}
High-precision assembly remains one of the most demanding problems in industrial robotics, as it requires aligning parts within submillimeter tolerances. Achieving such an alignment from vision is difficult: the gripper partially occludes the assembly interface, the surfaces are weakly textured and specular, and the interface itself covers only a few pixels of the image. A pose estimate acquired before the approach may be insufficient to complete the alignment. Visual feedback can correct the remaining displacement, but obtaining reliable feedback becomes challenging when the target itself is poorly observed.

Visual servoing controls robot motion by regulating image features or estimated poses relative to a desired configuration~\cite{chaumette2006visual, chaumette2007visual}. Object pose estimation instead aligns the robot with an explicit estimate of the target pose~\cite{foundationposewen2024}, a paradigm that still inherits the visibility limits described above. In contrast, scene-feature~\cite{vitvs2025, artvs2026} and direct~\cite{collewet2008visual, collewet2011photometric} methods regress motion from whatever the image contains. This broader visual context is useful in structured assembly cells: when a target remains fixed relative to its surroundings, the visible workspace structure can anchor a reference even when the target is partially occluded. Existing scene-based methods, however, reach only millimeter- to centimeter-level
accuracy or suffer from a small convergence basin, leaving open whether scene structure
can support the submillimeter accuracy that precise assembly demands.

Feed-forward visual geometry models offer a promising basis for this task. DUSt3R~\cite{dust3r2024} and Mast3R~\cite{mast3r2024} regress pointmaps from image pairs, and VGGT~\cite{vggt2025} and VGGT-$\Omega$~\cite{wang2026vggtomega} extend this to many views and additionally predict camera poses, in a single forward pass without per-scene optimization. These models regress camera motion from the
whole image rather than from the target alone, featuring
scene anchoring in exactly the sense required above. Their accuracy, however, is characterized only up to scale: reconstruction is evaluated after alignment to the ground truth, and relative-pose benchmarks score translation by direction, so neither establishes whether a predicted translation is millimeter-level accurate. Visual servoing instead requires both the direction of the remaining motion and its magnitude in physical units, and a scale that varies between frames makes a small stopping threshold impossible to apply. Closing this gap requires adapting the pretrained representation so that its translations become metric within the deployment workspace.

We present VGM-VS, a visual servoing framework that combines scene-anchored relative localization with scene-specific metric adaptation, as illustrated in Fig.~\ref{fig:teaser}. Given a current image and a reference image captured at the desired task configuration, VGM-VS estimates the relative camera pose from the surrounding scene structure, without an explicit estimate of the object pose. With the camera rigidly mounted on the robot end-effector, this estimate becomes the increment of a closed-loop controller, in which each new observation refines the previous command. To make the predicted translation metric, we fine-tune the camera head on data that the robot records autonomously in the scene. We evaluate VGM-VS on three real-world assembly tasks requiring precise positioning: USB-C cable picking, cable insertion, and RAM insertion, where it reaches submillimeter terminal accuracy on the cable tasks and tolerates large initial displacements, partial occlusions, and objects displaced during servoing.

Our contributions are summarized as follows:
\begin{itemize}

	\item \textbf{Scene-anchored visual servoing.}
	      A closed-loop frame-
	      work that uses static workspace geometry rather than
	      object models to servo to a reference image. Relative poses from a feed-forward visual geometry model guide iterative corrections, reducing the reliance on single-shot pose accuracy.

	\item \textbf{Scene-specific metric adaptation.} A procedure that fine-tunes the camera head on autonomously collected data while freezing the pretrained backbone, recovering metric pose predictions and the hand--eye transform without a dedicated calibration.

	\item \textbf{Real-robot validation.} An evaluation on three high-precision assembly tasks under large initial displacements, partial occlusions, and dynamic scene changes, showing lower terminal pose error and higher success rates than classical and learning-based baselines.

\end{itemize}

\section{Related Work}
\label{sec:related}

\subsection{Visual Servoing}
 
Classical visual servoing is divided into image-based visual servoing (IBVS), which minimizes feature errors in the image plane, and position-based visual servoing (PBVS), which minimizes a reconstructed pose error~\cite{hutchinson1996tutorial, chaumette2006visual, chaumette2007visual}; hybrid 2\,1/2-D schemes~\cite{malis1999} trade off between the local stability of IBVS and the dependence of PBVS on accurate pose estimation, calibration, and object models. Both families typically rely on sparse hand-crafted features such as SIFT~\cite{lowe2004sift}, ORB~\cite{rublee2011orb}, or AKAZE~\cite{alcantarilla2013akaze}, so precision degrades under weak texture, illumination changes, and occlusion, while direct photometric servoing~\cite{collewet2008visual, collewet2011photometric} avoids feature extraction altogether at the cost of a small convergence basin.
 
Learning-based methods relax these assumptions. Early works regress the relative camera pose with CNNs inside a PBVS loop~\cite{saxena2017exploring, bateux2018training}, reaching submillimeter accuracy with Siamese networks in scene-specific settings~\cite{yu2019siamese}, while others servo on learned representations such as optical flow~\cite{harish2020dfvs, katara2021deepmpcvs}, keypoints~\cite{puang2020kovis, adrian2022dfbvs}, or autoencoder and SE(3)-equivariant latent spaces~\cite{FeltonBFM22, Felton23a}. CNS~\cite{chen2024cns} encodes keypoint correspondences into a graph and trains a GNN controller purely in simulation, attaining submillimeter precision with a large convergence basin, and CNSv2~\cite{chen2025cnsv2} extends it to textureless scenes via probabilistic matching. ViT-VS~\cite{vitvs2025} and ART-VS~\cite{artvs2026} instead exploit pretrained Vision Transformer features within a classical IBVS loop, thereby generalizing without task-specific training. Still, existing pipelines either inherit the local stability and Jacobian conditioning issues of IBVS, or rely on pose estimators that lack a metric, geometry-grounded scene representation, limiting the achievable precision and robustness trade-off.
 
\subsection{Visual Geometry Models}
 
Feed-forward \emph{visual geometry models} replace multi-stage reconstruction pipelines with a single network. DUSt3R~\cite{dust3r2024} regresses dense pointmaps from uncalibrated image pairs, jointly encoding geometry, correspondence, and relative pose; MASt3R~\cite{mast3r2024} adds metric-scale outputs and dense descriptors, and follow-ups extend the paradigm to multi-view, streaming, and dynamic settings~\cite{spann3r2024, cut3r2025}. VGGT~\cite{vggt2025} unifies this family in one transformer that predicts cameras, depth, pointmaps, and point tracks for hundreds of views, $\pi^3$~\cite{pi32025} removes reference-view bias through permutation equivariance, and MapAnything~\cite{mapanything2025} handles heterogeneous input--output modalities at the metric scale. Most recently, VGGT-$\Omega$~\cite{wang2026vggtomega} shows that reconstruction quality scales predictably with model and data size, achieving state-of-the-art accuracy on static and dynamic scenes.
 
In robotics, these models serve as geometry-aware encoders for imitation policies~\cite{vggtdp2025, gp32025}, spatial-token providers for Vision Language Action~(VLA) models~\cite{omnivggt2025}, and front-ends for SLAM and relocalization~\cite{mast3rslam2025, relocvggt2025}. Yet they are evaluated almost exclusively on reconstruction and localization benchmarks where centimeter errors are tolerable. Whether such models can deliver the submillimeter, sub-degree accuracy demanded by closed-loop servoing remains open. VGM-VS addresses this gap by revisiting their geometric outputs as a metric feedback signal for high-precision servoing, benchmarked against classical IBVS~\cite{chaumette2006visual} and feature-driven ART-VS~\cite{artvs2026}.

\begin{figure*}[t]
	\centering
	\includegraphics[width=1.0\textwidth]{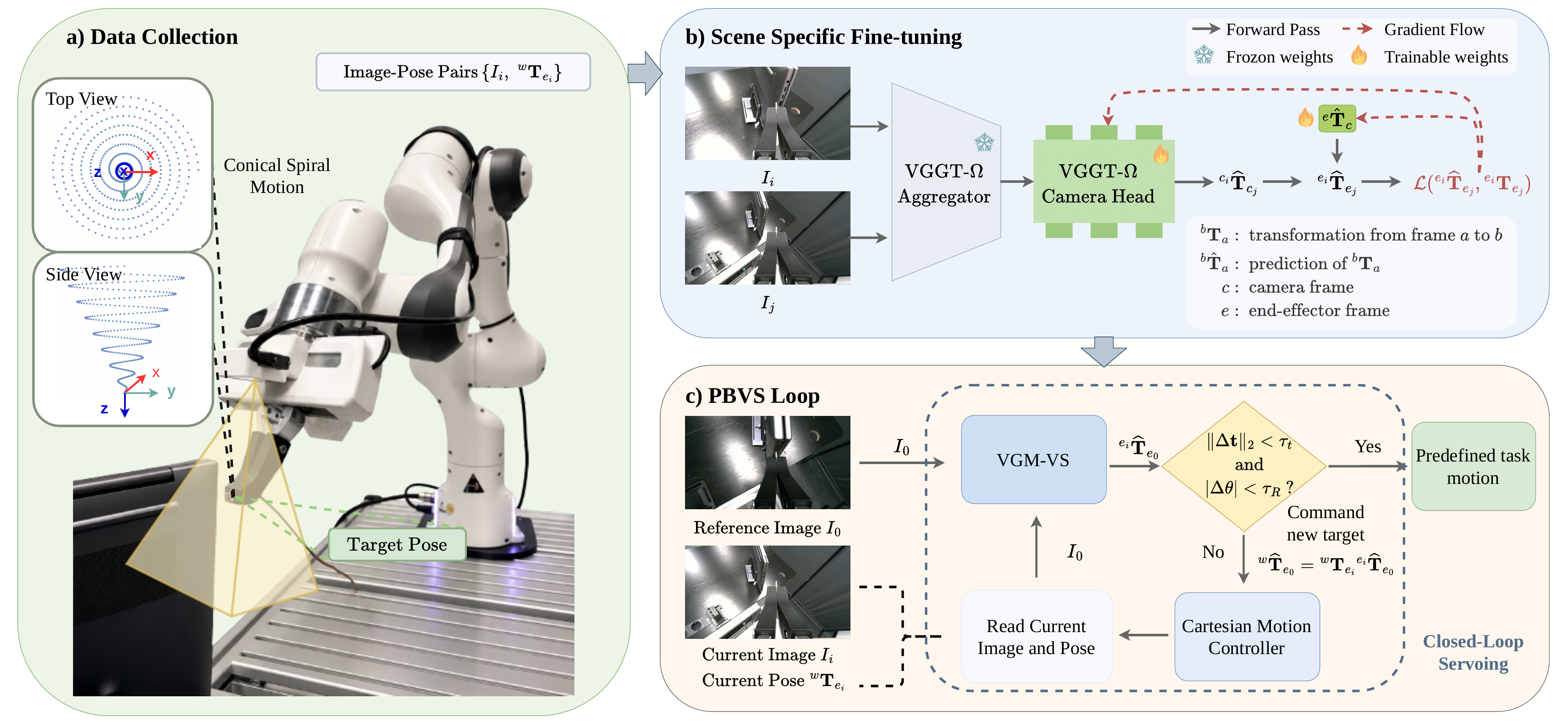}
    \vspace{-5mm}
	\caption{Overview of VGM-VS. (a) Data collection: the end-effector follows a conical spiral with its apex at the target pose, yielding synchronized image--pose pairs for the target workspace. (b) Fine-tuning: the aggregator is kept frozen, while the camera head and the hand--eye transform are trainable and optimized using the ground-truth relative pose \eqref{eq:gt_delta}. (c) Pose-based visual servoing (PBVS) loop: the predicted increment \eqref{eq:cam_to_ee} is composed with the current robot pose \eqref{eq:command} and sent to a Cartesian motion controller. When its translational and rotational magnitudes fall below $\tau_t$ and $\tau_R$, the predefined task motion is executed; otherwise, the loop continues with the new observation.}
	\label{fig:pipeline}
    \vspace{-5mm}
\end{figure*}

\section{Method}

In this section, we present VGM-VS, a scene-anchored visual servoing framework built on VGGT-$\Omega$~\cite{wang2026vggtomega} and summarized in Fig.~\ref{fig:pipeline}. VGM-VS positions the end-effector based on the current image $I$ and a reference image $I_0$, captured once at the desired end-effector pose with respect to a target object. We first describe how the camera predictions of VGGT-$\Omega$ are converted into a pose increment and used for closed-loop pose-based servoing control. We then address the scale ambiguity of these predictions in two ways. Sec.~\ref{sec:sim3} fits a global similarity calibration, which serves as a simple baseline, while Sec.~\ref{sec:finetuning} instead fine-tunes the model for metric-scale pose predictions, as adopted in VGM-VS. Finally, we introduce the automated procedure used to collect the scene-specific data on which both adaptations rely.

\subsection{Relative Pose Estimation with VGGT-$\Omega$}
\label{sec:relative_pose}

Given an input sequence of $(I_i)_{i=1}^N$ RGB images, where $I_i \in \mathbb{R}^{H \times W \times 3}$, VGGT-$\Omega$ jointly predicts the camera parameters and depth maps corresponding to each image in one forward pass~\cite{wang2026vggtomega}:

\begin{equation}
	f(I_1, \dots, I_N) = \big( (\mathbf{g}_1, D_1), \dots, (\mathbf{g}_N, D_N) \big)
	\label{eq:vggt_output}
\end{equation}

where $D_i \in \mathbb{R}^{H \times W}$ is the depth map of image $I_i$, and $\mathbf{g}_i = (\mathbf{q}_i, \mathbf{t}_i, \mathbf{f}_i) \in \mathbb{R}^{9}$ contains the rotation quaternion $\mathbf{q}_i \in \mathbb{R}^{4}$, the translation vector $\mathbf{t}_i \in \mathbb{R}^{3}$, and the field of view $\mathbf{f}_i \in \mathbb{R}^{2}$ of the corresponding camera. As in
VGGT~\cite{vggt2025}, all cameras are expressed in the frame of the first view, which is designated as the reference view using dedicated camera and register tokens.

We take the reference image as the first view, so that the frame of the model coincides with the camera frame $c_0$ at the target pose, and the prediction for the current view $I$ is directly the relative pose of the target with respect to the current camera, which the servoing loop of Sec.~\ref{sec:servoing} drives to the identity.

Internally, the aggregator $f_{\mathrm{agg}}$ maps the reference image $I_0$ and the current image $I$ to patch, camera, and scene tokens through DINOv3~\cite{dinov3} tokenization and alternating-attention with registers:
\begin{equation}
	f_{\mathrm{agg}}(I_0, I) = \big( (\mathbf{z}_0^{F'}, \mathbf{z}_0^{\mathrm{cam'}}, \mathbf{z}_0^{\mathrm{scene'}}), (\mathbf{z}^{F'}, \mathbf{z}^{\mathrm{cam'}}, \mathbf{z}^{\mathrm{scene'}}) \big).
	\label{eq:aggregator}
\end{equation}
Since depth prediction is not required for servoing settings, we drop the entire depth branch, evaluating only the camera head $f_{\mathrm{cam}}$, which jointly predicts the camera parameters for both views:
\begin{equation}
	f_{\mathrm{cam}}\big( (\mathbf{z}_0^{\mathrm{cam}\prime}, \mathbf{z}_0^{\mathrm{scene}\prime}), (\mathbf{z}^{\mathrm{cam}\prime}, \mathbf{z}^{\mathrm{scene}\prime}) \big) = \big( (\mathbf{q}_0, \mathbf{t}_0, \mathbf{f}_0), (\mathbf{q}, \mathbf{t}, \mathbf{f}) \big).
	\label{eq:camera_head}
\end{equation}
Here and below, ${}^{b}\mathbf{T}_{a} \in \mathrm{SE}(3)$ denotes the transformation of frame $a$ to frame $b$, and $\mathbf{T}(\mathbf{q}, \mathbf{t})$ the rigid transform assembled from a quaternion and a translation, and a hat denotes a quantity predicted by the model. As $c_0$ is the reference frame of the model, $\mathbf{g}_0$ is the identity up to regression error and is discarded, while $\mathbf{g} = (\mathbf{q}, \mathbf{t}, \mathbf{f})$ yields the relative pose
\begin{equation}
	{}^{c}\mathbf{T}_{c_0} = \mathbf{T}(\mathbf{q}, \mathbf{t}),
	\label{eq:relative_pose}
\end{equation}
which denotes the pose of the reference camera expressed in the current one. The scale ambiguity of the model affects only the norm of the predicted translation; its direction and the rotation remain valid.


The reference view thus serves as the visual anchor: the relative pose is estimated from the full field of view, so static workspace structure, such as fixtures, panels, and the table, provides localization cues when the target is small, textureless, or partially occluded. We refer to this formulation as \emph{scene-anchored} servoing, as it requires no object model, detector, or hand-crafted features.

\subsection{Closed-Loop Pose-Based Visual Servoing}
\label{sec:servoing}

In principle, the predicted relative pose can be used directly to guide the robot's motion towards the target. However the precision is the limiting factor, as it is bounded by that of a single prediction, made at an arbitrary, possibly large, distance from the target, where the object occupies only a few pixels in the image and the prediction is consequently noisier. Such a command also inherits the error of the hand--eye transform used to express it in the robot frame, and assumes a static scene throughout the motion. We therefore propose to apply this pose increment iteratively in a closed loop. Each successive iteration then moves the robot closer to the target, where predictions become gradually more accurate, analogously to a coarse-to-fine refinement, and small errors from the hand--eye calibration can also be compensated over the loop, as in IBVS~\cite{chaumette2006visual}. Another property of VGGT-$\Omega$ is that it is explicitly trained on dynamic scenes, which further improves the robustness to disturbances such as an occlusion or a displacement of part of the scene during servoing.

To express the predicted relative pose in the end-effector frame, we use the transformation ${}^{e}\mathbf{T}_{c}$, obtained through a conventional hand--eye calibration procedure:
\begin{equation}
	{}^{e}\mathbf{T}_{e_0} = {}^{e}\mathbf{T}_{c} \, {}^{c}\mathbf{T}_{c_0} \, \big( {}^{e}\mathbf{T}_{c} \big)^{-1},
	\label{eq:cam_to_ee}
\end{equation}
and compose it with the current robot pose to obtain the pose commanded to a Cartesian motion controller:
\begin{equation}
	{}^{w}\mathbf{T}_{e_0} = {}^{w}\mathbf{T}_{e} \, {}^{e}\mathbf{T}_{e_0}.
	\label{eq:command}
\end{equation}
In the loop, we acquire the next image without waiting for the motion to complete, so that each prediction overrides the previous command. Since the predicted direction is correct even at large distances, this sequence leads to a smooth, direct motion towards the target pose, along which the predicted increments decay monotonically to zero. The convergence is then determined by the translational and rotational magnitudes of ${}^{e}\mathbf{T}_{e_0}$ that fall below the thresholds $\tau_t$ and $\tau_R$, which are selected according to the tolerance required by the task. In manipulation tasks, the reference image is usually captured at a pre-grasp or pre-insertion pose, so that after convergence, the task is completed by a short predefined motion in task space. The whole process is illustrated in Fig.~\ref{fig:pipeline}(c).

\subsection{Global Similarity Calibration}
\label{sec:sim3}

VGGT-$\Omega$~\cite{wang2026vggtomega} is trained with normalized scene supervision. Its predictions are thus recovered only up to an unknown global scale and cannot be used directly for robot control.

To resolve this scale ambiguity, we first consider a scene-specific $\mathrm{Sim}(3)$ correction, fitted with the ground-truth relative poses \eqref{eq:gt_delta} of the scene-specific data as later described in Sec.~\ref{sec:data_collection}, which we express in the camera frame with the calibrated ${}^{e}\mathbf{T}_{c}$. Since the predicted and the ground-truth translations are relative to the same reference view, the correction preserves the origin and involves no translation term, so the transformation reduces to a scale $s > 0$ and a rotation $\mathbf{R} \in \mathrm{SO}(3)$. We estimate both by formulating a least-squares problem over the $K$ collected samples, denoting by $(\hat{\mathbf{t}}_k, \hat{\mathbf{q}}_k)$ and $(\mathbf{t}_k, \mathbf{q}_k)$ the translation and the unit quaternion part of the predicted and ground-truth relative pose of the $k$-th sample:
\begin{equation}
	s^{\star}, \mathbf{R}^{\star} = \operatorname*{arg\,min}_{s > 0, \; \mathbf{R} \in \mathrm{SO}(3)} \sum_{k=1}^{K} \big\| s \, \mathbf{R} \, \hat{\mathbf{t}}_k - \mathbf{t}_k \big\|^{2},
	\label{eq:sim3_objective}
\end{equation}
which can be solved in closed form with Umeyama alignment~\cite{umeyama1991least}. With $\sum_{k} \hat{\mathbf{t}}_k \mathbf{t}_k^{\top} = \mathbf{U} \boldsymbol{\Sigma} \mathbf{V}^{\top}$ the singular value decomposition (SVD) and $\varepsilon > 0$ a small constant for numerical stability, the solution is as follows:
\begin{equation}
	\begin{split}
		\mathbf{R}^{\star} & = \mathbf{V} \, \mathrm{diag}\big(1, 1, \det(\mathbf{V}\mathbf{U}^{\top})\big) \, \mathbf{U}^{\top},                                                   \\[2pt]
		s^{\star}          & = \frac{\sum_{k} \big\langle \mathbf{R}^{\star} \hat{\mathbf{t}}_k, \, \mathbf{t}_k \big\rangle}{\sum_{k} \| \hat{\mathbf{t}}_k \|^{2} + \varepsilon},
	\end{split}
	\label{eq:sim3_solution}
\end{equation}
The fitted parameters are then applied to every prediction of \eqref{eq:relative_pose} before it enters the control loop: the translation is mapped to $s^{\star} \mathbf{R}^{\star} \hat{\mathbf{t}}$, while the predicted rotation is expressed in the corrected frame as $\mathbf{R}^{\star} \hat{\mathbf{R}} \mathbf{R}^{\star\top}$, so that the scale only affects the translation. We refer to this variant as VGM-VS w/o FT (without fine-tuning).

\subsection{Camera Head Fine-Tuning}
\label{sec:finetuning}

VGGT-$\Omega$ was not trained for the level of accuracy required by high-precision tasks, and its residual error varies with the viewpoint and the distance to the target. A single global similarity transformation thus cannot compensate for such an error, as later seen in Sec.~\ref{sec:experiments}. Rather than correcting the model's output, we therefore let the model adapt to the scene itself.

Starting from the pretrained weights, we keep the aggregator \eqref{eq:aggregator} frozen and fine-tune the camera head \eqref{eq:camera_head}, which can then learn the non-linear mapping from the aggregated tokens to metric poses. Freezing the aggregator preserves the general geometric representation learned during large-scale pretraining, which a few thousand images of a single scene could otherwise degrade.

The training samples share the structure described in Sec.~\ref{sec:data_collection}, in which the reference image $I_0$ is paired with a recorded image $I_i$. With probability $p_{\mathrm{id}}$, the latter is instead paired with itself, with the relative pose \eqref{eq:gt_delta} set to identity. Such samples emphasize the predictions near the target pose, where noise leads to oscillations, slow convergence and a larger terminal error. Otherwise, with probability $p_{\mathrm{rand}}$, the reference is replaced with a randomly drawn pair $(I_j, {}^{w}\mathbf{T}_{e_j})$, $j \neq i$. This increases the variance of the viewpoints seen during training and avoids a drop in performance when a different reference image is used at inference, as is common in practice. To increase the generalizability of the camera head predictions, we further augment the images with random brightness, contrast, saturation, and gamma changes, followed by Gaussian blur, for robustness to the illumination changes and motion blur which are commonly encountered at deployment.

The predicted end-effector increment \eqref{eq:cam_to_ee} is supervised against its ground-truth counterpart \eqref{eq:gt_delta} by a combination of translational and rotational loss terms:
\begin{equation}
	\mathcal{L} = \lambda_{t} \, \mathcal{L}_{t} + \lambda_{R} \, \mathcal{L}_{R},
	\label{eq:loss_total}
\end{equation}
both evaluated over a minibatch of $B$ samples. The translation is penalized by the standard mean squared error:
\begin{equation}
	\mathcal{L}_{t} = \frac{1}{B} \sum_{k=1}^{B} \frac{\big\| \hat{\mathbf{t}}_k - \mathbf{t}_k \big\|^{2}}{d_{\max}^{2}},
	\label{eq:loss_translation}
\end{equation}
where $d_{\max} > 0$ is a reference distance that renders the term dimensionless. The rotation is penalized by the geodesic angle between the ground-truth and the predicted unit quaternion:
\begin{equation}
	\mathcal{L}_{R} = \frac{1}{B} \sum_{k=1}^{B} \frac{\arccos\big( 2 \langle \hat{\mathbf{q}}_k, \mathbf{q}_k \rangle^{2} - 1 \big)}{\theta_{\max}},
	\label{eq:loss_rotation}
\end{equation}
where the inner product is squared to make the loss invariant to the sign ambiguity of the quaternion double cover, and $\theta_{\max} > 0$ normalizes the angle in the same way.

Translation and rotation are expressed in different units; balancing their loss terms is a known challenge in pose regression~\cite{kendall17_pose_regression}. We handle it with a simple set of hyperparameters: the normalization by $d_{\max}$ and $\theta_{\max}$ makes both terms dimensionless and comparable, while $\lambda_{t}$ and $\lambda_{R}$ set their relative contribution.

Notably, the hand--eye transform no longer requires a dedicated calibration procedure: initialized at the identity, a learned transform ${}^{e}\hat{\mathbf{T}}_{c}$ replaces ${}^{e}\mathbf{T}_{c}$ in \eqref{eq:cam_to_ee} and is optimized directly from the loss supervision, using the data collected in the scene.

\subsection{Data Collection}
\label{sec:data_collection}

Both approaches presented in Secs.~\ref{sec:sim3} and~\ref{sec:finetuning} rely on real-world data of the scene, which are
collected automatically: starting from the target pose, the
end-effector follows a precomputed trajectory that sweeps the free space
region around it, in practice, along the direction opposite to the predefined
task motion of Sec.~\ref{sec:servoing}. The objective of this motion is to
capture as much visual variance as possible, i.e.\ a wide range of viewpoints
and translational distances, in a short recording time, while
concentrating the samples near the target pose, where the prediction
accuracy matters most.

For that, we use a conical spiral with its apex at the target pose, parameterized by
its maximum radius $r_{\max}$, its length $\ell$ along the approach axis, and
its number of turns $n$, as illustrated in Fig.~\ref{fig:pipeline}(a); other
sweeping patterns, such as a pyramidal one, would be equally valid. At each
waypoint the camera is oriented towards the target object to guarantee its
visibility, and a slight random perturbation is added to the resulting
rotation.

Since the trajectory is executed without stopping at the
waypoints, the robot is still moving when an image is acquired, and any
temporal misalignment between the image and robot-state streams translates
directly into label noise. To alleviate this, we timestamp each image at the
midpoint of its exposure interval and pair it with the pose
obtained by interpolating between the two closest robot states, with both streams sharing a common clock. This yields a set of synchronized pairs
$(I_i, {}^{w}\mathbf{T}_{e_i})$ of images and end-effector poses. In practice,
the recording process completes within a few minutes, producing a dataset of
several thousand pairs, which we empirically find sufficient to adapt the
model to a new scene, as shown in Sec.~\ref{sec:experiments}.

Each training sample is then formed by pairing the reference image $I_0$, recorded at the target pose, with another image $I_i$, $i \neq 0$, and labeled with the relative transform in end-effector frame
\begin{equation}
	{}^{e_i}\mathbf{T}_{e_0} = \big( {}^{w}\mathbf{T}_{e_i} \big)^{-1} \, {}^{w}\mathbf{T}_{e_0},
	\label{eq:gt_delta}
\end{equation}
which is the ground-truth counterpart of \eqref{eq:cam_to_ee}.

\section{Experiments}
\label{sec:experiments}

\begin{table*}[!t]
	\centering
	\caption{Quantitative comparison on three real-world tasks. $\uparrow$/$\downarrow$ indicate that higher/lower is better; the best result within each setting is shown in \textbf{bold}, and a dash denotes a metric that is not measured.}
    \vspace{-3mm}
	\label{tab:experimental_results}
	\renewcommand{\arraystretch}{1.0}
	\setlength{\tabcolsep}{2pt}
	\begin{tabular}{lll l c c c c c}
		\toprule
		Task & Eval. type                        & Setting                    & Method        & Conv. (\%) $\uparrow$ & $e_t$ (mm) $\downarrow$ & $e_R$ ($^\circ$) $\downarrow$ & Iter. $\downarrow$ & Success (\%) $\uparrow$ \\
		\midrule
		\multirow{20}{*}{RAM insertion}
		     & \multirow{8}{*}{Init. pose pert.}
		     & \multirow{4}{*}{Small}
		     & ORB-IBVS                          & \textbf{100}               & 3.45          & 0.8                   & 777.34                  & --                                                                           \\
		     &                                   &                            & ART-VS        & 46                    & 32.97                   & 6.47                          & 140.11             & --                      \\
		     &                                   &                            & VGM-VS w/o FT & \textbf{100}          & 6.6                     & 1.04                          & \textbf{23}        & --                      \\
		     &                                   &                            & VGM-VS        & \textbf{100}          & \textbf{0.95}           & \textbf{0.45}                 & 26.64              & --                      \\
		\cmidrule(lr){3-9}
		     &                                   & \multirow{4}{*}{Large}
		     & ORB-IBVS                          & 66                         & 3.69          & 0.78                  & 926.33                  & --                                                                           \\
		     &                                   &                            & ART-VS        & 27                    & 65.34                   & 15.11                         & 110.35             & --                      \\
		     &                                   &                            & VGM-VS w/o FT & 96                    & 6.95                    & 1.11                          & 29.75              & --                      \\
		     &                                   &                            & VGM-VS        & \textbf{100}          & \textbf{1.15}           & \textbf{0.55}                 & \textbf{26.79}     & --                      \\
		\cmidrule(lr){2-9}
		     & \multirow{12}{*}{Scene dist.}
		     & \multirow{4}{*}{25\% occ.}
		     & ORB-IBVS                          & 40                         & 5.43          & 1.15                  & 1101                    & --                                                                           \\
		     &                                   &                            & ART-VS        & 65                    & 30.12                   & 6.20                          & 79                 & --                      \\
		     &                                   &                            & VGM-VS w/o FT & \textbf{100}          & 5.55                    & 1.25                          & 21                 & --                      \\
		     &                                   &                            & VGM-VS        & \textbf{100}          & \textbf{3.37}           & \textbf{0.87}                 & \textbf{17.65}     & --                      \\
		\cmidrule(lr){3-9}
		     &                                   & \multirow{4}{*}{50\% occ.}
		     & ORB-IBVS                          & 15                         & 7.32          & 1.66                  & 1297                    & --                                                                           \\
		     &                                   &                            & ART-VS        & 55                    & 29.82                   & 5.84                          & 41.18              & --                      \\
		     &                                   &                            & VGM-VS w/o FT & 85                    & 14.8                    & 2.85                          & 43.29              & --                      \\
		     &                                   &                            & VGM-VS        & \textbf{100}          & \textbf{4.21}           & \textbf{1.07}                 & \textbf{19.4}      & --                      \\
		\cmidrule(lr){3-9}
		     &                                   & \multirow{4}{*}{Dynamic}
		     & ORB-IBVS                          & 55                         & --            & --                    & 1951                    & 30                                                                           \\
		     &                                   &                            & ART-VS        & 20                    & --                      & --                            & 39.5               & 10                      \\
		     &                                   &                            & VGM-VS w/o FT & 90                    & --                      & --                            & 27                 & 60                      \\
		     &                                   &                            & VGM-VS        & \textbf{100}          & --                      & --                            & \textbf{20.2}      & \textbf{95}             \\
		\midrule
		\multirow{4}{*}{Cable picking}
		     & \multirow{2}{*}{Init. pose pert.}
		     & \multirow{2}{*}{Large}
		     & VGM-VS w/o FT                     & \textbf{100}               & 4.61          & 0.32                  & 36                      & --                                                                           \\
		     &                                   &                            & VGM-VS        & \textbf{100}          & \textbf{0.58}           & \textbf{0.20}                 & \textbf{26.56}     & --                      \\
		\cmidrule(lr){2-9}
		     & \multirow{2}{*}{Scene dist.}
		     & \multirow{2}{*}{Dynamic}
		     & VGM-VS w/o FT                     & 90                         & --            & --                    & 17.38                   & 10                                                                           \\
		     &                                   &                            & VGM-VS        & \textbf{100}          & --                      & --                            & \textbf{12.29}     & \textbf{100}            \\
		\midrule
		\multirow{4}{*}{Cable insertion}
		     & \multirow{2}{*}{Init. pose pert.}
		     & \multirow{2}{*}{Large}
		     & VGM-VS w/o FT                     & \textbf{100}               & 1.98          & 0.33                  & \textbf{26.62}          & --                                                                           \\
		     &                                   &                            & VGM-VS        & \textbf{100}          & \textbf{0.67}           & \textbf{0.08}                 & 33.06              & --                      \\
		\cmidrule(lr){2-9}
		     & \multirow{2}{*}{Scene dist.}
		     & \multirow{2}{*}{Dynamic}
		     & VGM-VS w/o FT                     & 95                         & --            & --                    & \textbf{11.6}           & 0                                                                            \\
		     &                                   &                            & VGM-VS        & \textbf{100}          & --                      & --                            & 15                 & \textbf{90}             \\
		\bottomrule
	\end{tabular}
    \vspace{-5mm}
\end{table*}

\subsection{Compared Methods}
We compare VGM-VS with ORB-IBVS, a classical baseline using ORB keypoints and descriptors, and ART-VS, a state-of-the-art learned-feature visual servoing method, for which we use the official implementation. Both baselines compute a camera-frame twist with the point-feature IBVS control law. To use the same downstream Cartesian motion controller for all methods, we integrate each twist over the control period $\Delta t = 1/f$ on $\mathrm{SE}(3)$ and express the resulting pose increment in the end-effector frame using ${}^{e}\mathbf{T}_{c}$.

As an ablation, we evaluate VGM-VS w/o FT, the globally calibrated variant of Sec.~\ref{sec:sim3}, to isolate the contribution of scene-specific fine-tuning. Its $\mathrm{Sim}(3)$ correction is fitted on a subset of the same data used for fine-tuning.

\subsection{Experimental Setup}
\paragraph{Tasks}
We evaluate the methods on three high-precision industrial manipulation tasks: USB-C cable picking, USB-C cable insertion, and RAM insertion. The cable tasks use a Franka Emika Panda robot equipped with a Franka Hand and a wrist-mounted RealSense D405 camera. The robot grasps a cable from a holder and inserts it into a laptop USB-C port. The smooth, untextured surfaces of this scene provide too few distinctive features for ORB-IBVS and ART-VS to establish the four correspondences required by the IBVS control law, so only VGM-VS and VGM-VS w/o FT are evaluated on these two tasks. RAM insertion uses an Agile Robots Thor 3 robot equipped with a Robotiq gripper and the same wrist-mounted camera. The motherboard's dense arrangement of electronic components provides sufficient visual features, allowing all four methods to be evaluated.

\paragraph{Implementation details}
For each scene, we collect approximately $10{,}000$ image--pose pairs as described in Sec.~\ref{sec:data_collection}, over a two-minute trajectory with the camera running at 90\,Hz. The conical spiral is parameterized by a maximum radius $r_{\max}=10$\,cm, a length $\ell=30$\,cm along the approach axis, and $n=8$ turns, and the look-at orientation of each waypoint is perturbed by up to $1^\circ$. Starting from the pretrained VGGT-$\Omega$ checkpoint, we fine-tune VGM-VS for 200 epochs with a batch size of 64, using AdamW with a linear warm-up over the first 20 epochs followed by a cosine decay, and sampling the training pairs with $p_{\mathrm{id}}=0.01$ and $p_{\mathrm{rand}}=0.5$. Fine-tuning takes approximately two hours on eight NVIDIA RTX~6000 Pro GPUs. During evaluation, all methods run on the workstation equipped with an AMD Ryzen~9 9950X CPU and a NVIDIA RTX~5090 GPU, and the loop is considered converged once the commanded increment falls below $\tau_t = 0.3$\,mm and $\tau_R = 0.3^\circ$. VGM-VS runs at 30\,Hz with $480\times640$ images. ART-VS and ORB-IBVS achieve raw inference rates of 132\,Hz and 110\,Hz, respectively, using $720\times1280$ images; however, their closed-loop control frequency is capped at 30\,Hz by the RealSense D405 camera’s maximum frame rate at this resolution.

\subsection{Evaluation}
\paragraph{Test conditions}
We evaluate robustness to initial-pose perturbations and scene disturbances. Initial-pose perturbations are generated relative to the camera pose at which the reference image was captured. We first apply a translational offset and orient the camera toward the target object to ensure visibility; the rotational perturbation is then applied relative to this look-at orientation. A perturbation is classified as small when its translational and rotational
magnitudes do not exceed 15\,cm and $15^\circ$, respectively, and as large
otherwise, reaching up to 30\,cm and $30^\circ$. Scene disturbances include occluding 25\% or 50\% of the target object and a dynamic setting in which the object is moved several centimeters after the reference image is captured. We conduct 100 trials for each initial-pose perturbation setting and 20 trials for each scene-disturbance setting, using the same reference image within each scene. Convergence and task-success rates are computed over all trials, while pose errors and iteration counts are averaged over converged trials.

\newlength{\plotwide}
\newlength{\plotnarrow}
\setlength{\plotwide}{0.332\textwidth}
\setlength{\plotnarrow}{0.332\textwidth}

\newcommand{\plotpanel}[3]{%
	\begin{minipage}[t]{#1}%
		\centering
		\includegraphics[width=\linewidth]{#2}%
		\par\vspace{-2pt}{\scriptsize (#3)}%
	\end{minipage}%
}

\begin{figure*}[!t]
	\centering
	\plotpanel{\plotnarrow}{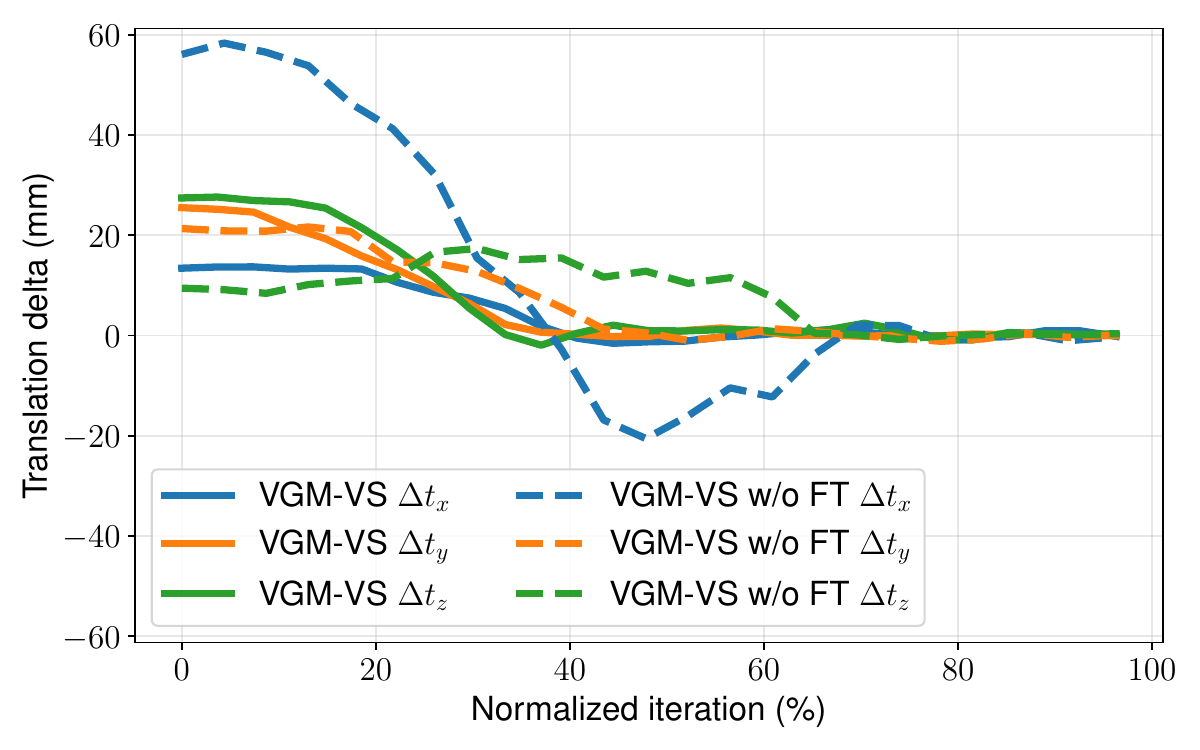}{a}\hfill
	\plotpanel{\plotnarrow}{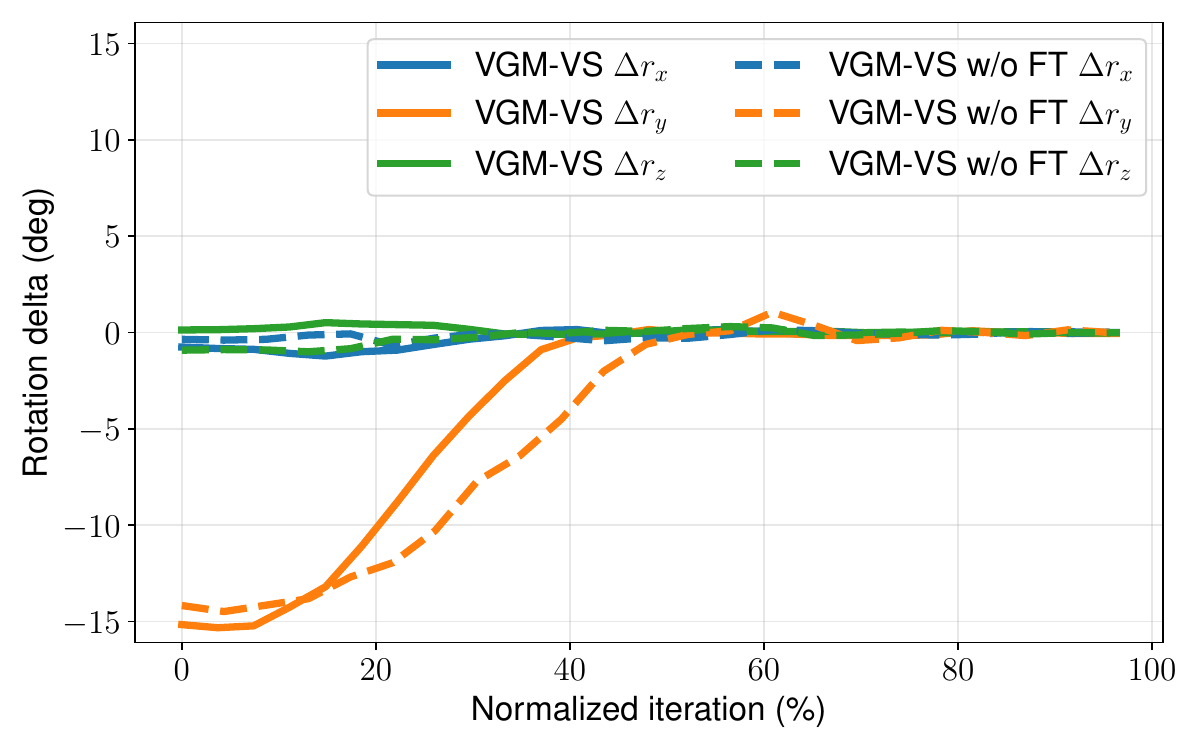}{b}\hfill
	\plotpanel{\plotwide}{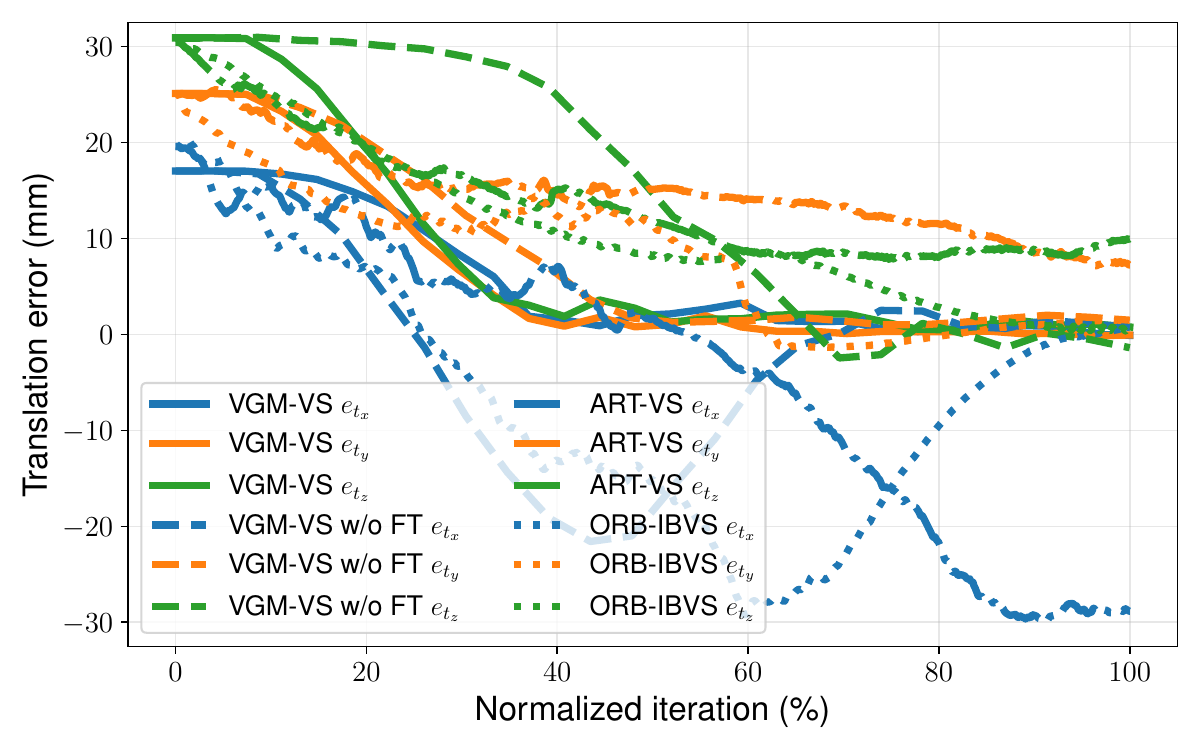}{c}\par\vspace{4pt}
	\plotpanel{\plotnarrow}{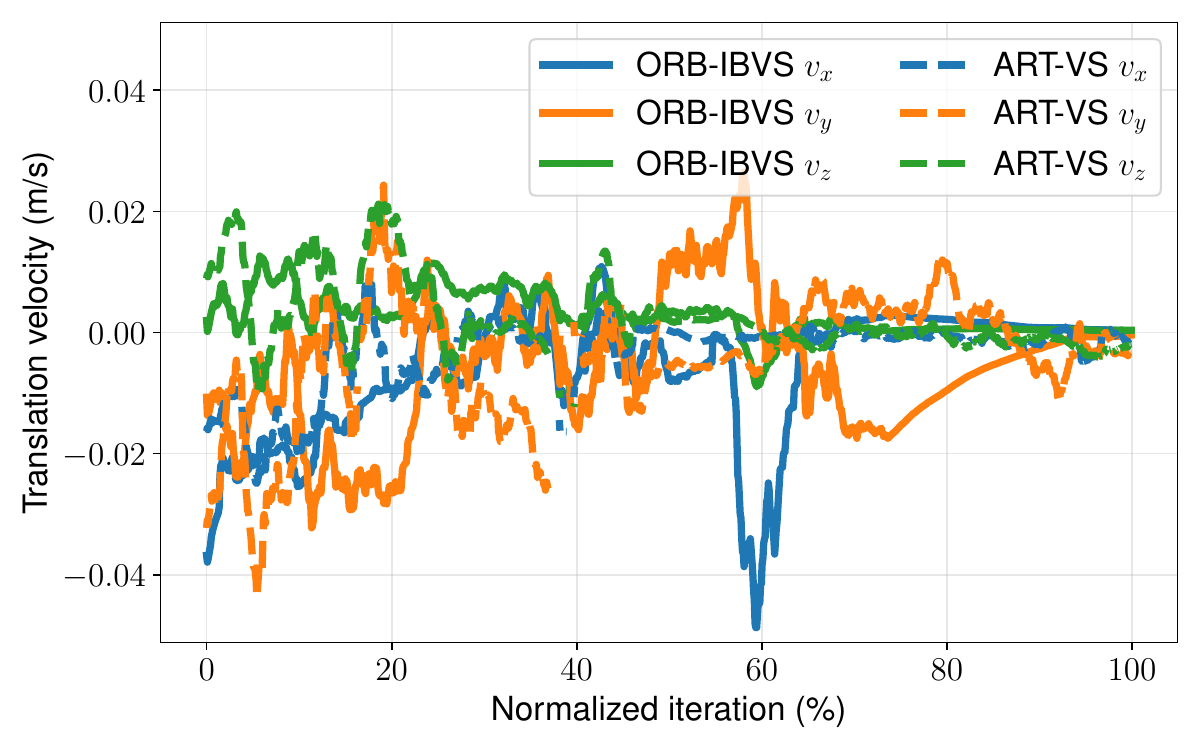}{d}\hfill
	\plotpanel{\plotnarrow}{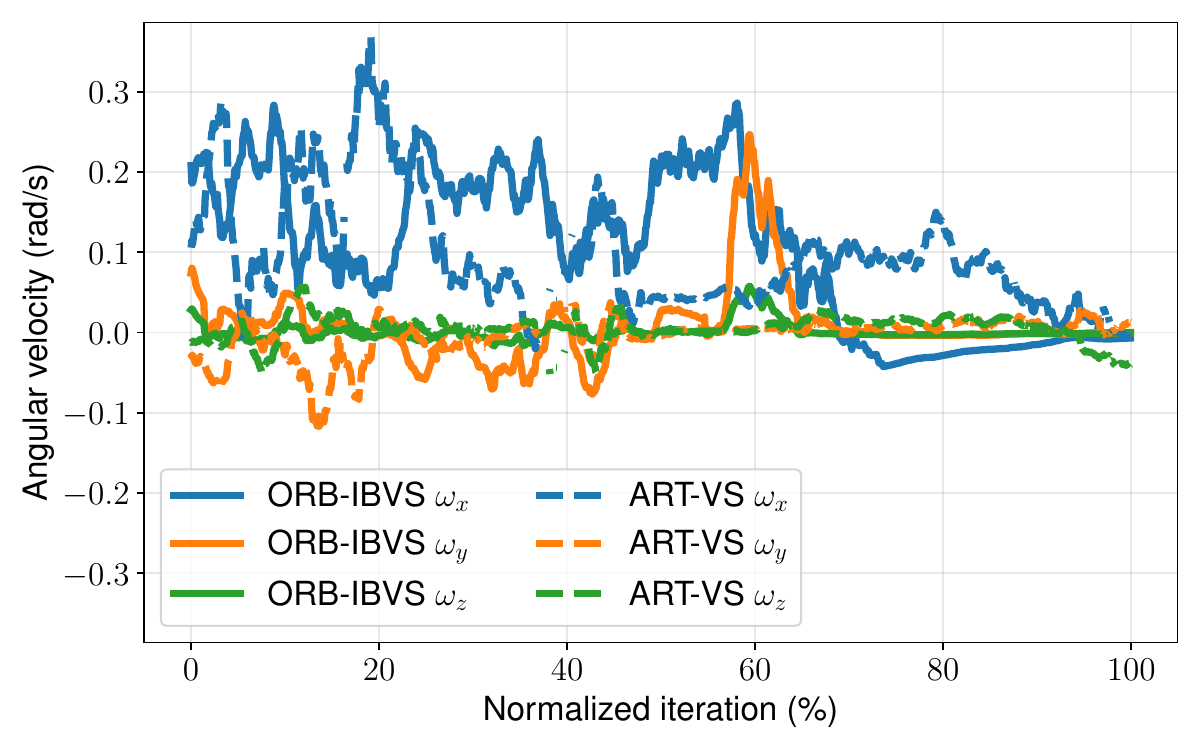}{e}\hfill
	\plotpanel{\plotwide}{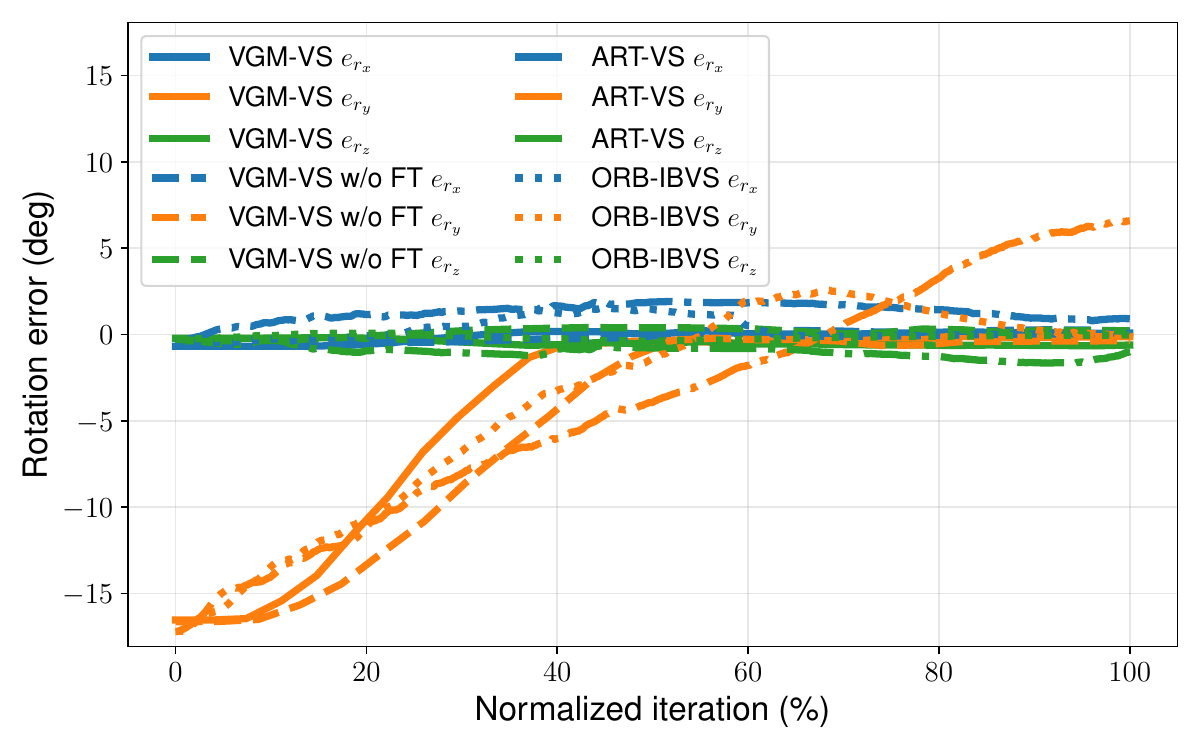}{f}
    \vspace{-3mm}
	\caption{RAM insertion under large initial pose perturbation. The $x$ axis represents the number of inference steps (\%); the $y$ axis is (a) predicted translation increment $\Delta t$ (mm); (b) predicted rotation increment $\Delta r$ (deg); (c) translation error $e_t$ (mm); (d) camera linear velocity $v$ (m/s); (e) camera angular velocity $\omega$ (rad/s); (f) rotation error $e_R$ (deg). VGM-VS variants are shown in (a), (b); the keypoint-based baselines in (d), (e); all four methods in (c), (f).}
	\label{fig:ram_insertion_plots}
    \vspace{-5mm} 
\end{figure*}

\paragraph{Metrics}
For static settings, we report the convergence rate, final translation error $e_t$, final rotation error $e_R$, and the number of inference iterations required for convergence. Pose errors are measured relative to the ground-truth pose at which the reference image was captured. In the dynamic setting, where the target pose is unavailable, we report convergence rate, iteration count, and task-success rate. A trial is classified as non-convergent if a collision occurs, an unreachable pose is commanded, or convergence is not achieved within one minute.

\subsection{Quantitative Results}
Table~\ref{tab:experimental_results} shows that VGM-VS converges in every setting of every task. The cable tasks separate the methods most sharply: the keypoint-based baselines are inapplicable there, whereas VGM-VS converges by anchoring on the target's surroundings. On RAM insertion, where all four methods can be compared, the baselines degrade as the conditions get harder: under 50\% occlusion, the convergence rates of ORB-IBVS, ART-VS, and VGM-VS w/o FT fall to 15\%, 55\%, and 85\%. ORB-IBVS establishes more correspondences than ART-VS, which likely explains its higher convergence rate under pose perturbations, but its less distinctive features produce more outliers and a slow, oscillatory behavior costing an order of magnitude more iterations than VGM-VS: in Fig.~\ref{fig:ram_insertion_plots}(d) and (e), the velocities of both baselines keep fluctuating along the trajectory, whereas those of VGM-VS decay smoothly to zero. ART-VS instead matches fewer points, so a few wrong ones bias the feature error and drive the camera to a wrong pose reported as converged, explaining its low iteration count and the large terminal errors in Fig.~\ref{fig:ram_insertion_plots}(c) and (f).

Scene-specific fine-tuning turns coarse alignment into task-level precision. VGM-VS w/o FT aligns the camera reliably in most trials, but its predictions remain inaccurate: in Fig.~\ref{fig:ram_insertion_plots}(a) and (b) its increments overshoot and change sign before settling, unlike those of the fine-tuned model, which decay monotonically to zero. Since the loop terminates once the increment falls below $\tau_t$ and $\tau_R$, such noise causes premature convergence several millimeters away from the target: in dynamic cable insertion, it stops after $11.6$ iterations yet succeeds in none of the trials. Fine-tuning instead yields the best accuracy in every setting, with submillimeter translation and sub-degree rotation errors on both cable tasks, and drives both errors close to zero in Fig.~\ref{fig:ram_insertion_plots}(c) and (f). Such precision is what the tasks require, as each ends with a predefined linear motion along a single axis: VGM-VS reaches task-success rates of 95\%, 100\%, and 90\% for dynamic RAM insertion, cable picking, and cable insertion, against 60\%, 10\%, and 0\% for VGM-VS w/o FT.

\section{Conclusion}

In this work we have introduced VGM-VS, a scene-anchored visual servoing framework that uses a pretrained visual geometry model to estimate the relative pose between the current and the reference image, and applies it as the pose increment for a closed-loop controller. No object model, detector, or feature correspondences are needed: a single reference image defines the goal, and the surrounding workspace provides the cues when the target itself is barely visible. A short, autonomously collected robot dataset is used to fine-tune the camera head and jointly refine the hand--eye transformation, recovering metric scale and reducing the viewpoint-dependent errors that a global similarity transformation cannot correct. Adapting a new scene thus costs two minutes of recording and about two hours of training, without human intervention. Across experiments on cable picking, cable insertion, and RAM insertion, VGM-VS converged in all evaluated settings while running at 30\,Hz, reaching submillimeter terminal error and sub-degree rotations on the two cable tasks, and 90--100\% task success when the object is displaced during servoing, while remaining robust to texture-poor scenes, large initial displacements, and partial occlusions.

The main limitation is that the adaptation remains scene specific, as a new workspace requires another recording and training session; generalizing a single adapted model to unseen scenes is therefore an important direction for future work. Longer-horizon tasks would also require an automatic selection of the intermediate reference views, since specifying one per alignment stage by hand would introduce substantial task-specific engineering. Finally, the short contact-rich motion that completes each task is executed open loop, and integrating force feedback during this phase could further improve the task success rate.


\section*{ACKNOWLEDGMENT}

The authors thank Xuming Meng for reviewing the manuscript,
Zewen Yang for helpful discussions, and Swapnil Mule and his team
for the fixture design and 3D printing used in the experiments.
This work was carried out at Agile Robots SE.



\bibliographystyle{IEEEtran}
\bibliography{references}

@IEEEtranBSTCTL{IEEEexample:BSTcontrol,
  CTLdash_repeated_names = "no"
}

@article{hutchinson1996tutorial,
  author  = {Hutchinson, Seth and Hager, Gregory D. and Corke, Peter I.},
  title   = {A Tutorial on Visual Servo Control},
  journal = {IEEE Transactions on Robotics and Automation},
  volume  = {12},
  number  = {5},
  pages   = {651--670},
  year    = {1996}
}

@article{chaumette2006visual,
  author  = {Chaumette, Fran{\c{c}}ois and Hutchinson, Seth},
  title   = {Visual Servo Control. {I}. {B}asic Approaches},
  journal = {IEEE Robotics \& Automation Magazine},
  volume  = {13},
  number  = {4},
  pages   = {82--90},
  year    = {2006}
}

@article{chaumette2007visual,
  author  = {Chaumette, Fran{\c{c}}ois and Hutchinson, Seth},
  title   = {Visual Servo Control. {II}. {A}dvanced Approaches},
  journal = {IEEE Robotics \& Automation Magazine},
  volume  = {14},
  number  = {1},
  pages   = {109--118},
  year    = {2007}
}

@article{malis1999,
  author  = {Malis, Ezio and Chaumette, Fran{\c{c}}ois and Boudet, Sylvie},
  title   = {2 1/2 {D} Visual Servoing},
  journal = {IEEE Transactions on Robotics and Automation},
  volume  = {15},
  number  = {2},
  pages   = {238--250},
  year    = {1999}
}

@article{lowe2004sift,
  author  = {Lowe, David G.},
  title   = {Distinctive Image Features from Scale-Invariant Keypoints},
  journal = {International Journal of Computer Vision},
  volume  = {60},
  number  = {2},
  pages   = {91--110},
  year    = {2004}
}

@inproceedings{rublee2011orb,
  author    = {Rublee, Ethan and Rabaud, Vincent and Konolige, Kurt and Bradski, Gary},
  title     = {{ORB}: An Efficient Alternative to {SIFT} or {SURF}},
  booktitle = {IEEE International Conference on Computer Vision (ICCV)},
  pages     = {2564--2571},
  year      = {2011}
}

@inproceedings{alcantarilla2013akaze,
  author    = {Alcantarilla, Pablo F. and Nuevo, Jes{\'u}s and Bartoli, Adrien},
  title     = {Fast Explicit Diffusion for Accelerated Features in Nonlinear Scale Spaces},
  booktitle = {British Machine Vision Conference (BMVC)},
  year      = {2013}
}

@article{collewet2011photometric,
  author  = {Collewet, Christophe and Marchand, Eric},
  title   = {Photometric Visual Servoing},
  journal = {IEEE Transactions on Robotics},
  volume  = {27},
  number  = {4},
  pages   = {828--834},
  year    = {2011}
}

@inproceedings{collewet2008visual,
  title        = {Visual servoing set free from image processing},
  author       = {Collewet, Christophe and Marchand, Eric and Chaumette, Fran{\c{c}}ois},
  booktitle    = {2008 IEEE International Conference on Robotics and Automation (ICRA)},
  pages        = {21--26},
  year         = {2008},
  organization = {IEEE}
}

@inproceedings{Felton23a,
  author    = {Felton, S. and Fromont, E. and Marchand, E.},
  title     = {Deep metric learning for visual servoing: when pose and image meet in latent space},
  booktitle = {{IEEE Int. Conf. on Robotics and Automation, ICRA'23}},
  pages     = {741--747},
  address   = {London, UK},
  month     = {May},
  year      = {2023}
}

@article{FeltonBFM22,
  author    = {Samuel Felton and
               Pascal Brault and
               {\'{E}}lisa Fromont and
               {\'{E}}ric Marchand},
  title     = {Visual Servoing in Autoencoder Latent Space},
  journal   = {{IEEE} Robotics Automation Lett.},
  volume    = {7},
  number    = {2},
  pages     = {3234--3241},
  year      = {2022}
}

@misc{foundationposewen2024,
  title         = {FoundationPose: Unified 6D Pose Estimation and Tracking of Novel Objects},
  author        = {Bowen Wen and Wei Yang and Jan Kautz and Stan Birchfield},
  year          = {2024},
  eprint        = {2312.08344},
  archiveprefix = {arXiv},
  primaryclass  = {cs.CV},
  url           = {https://arxiv.org/abs/2312.08344}
}

@inproceedings{saxena2017exploring,
  author    = {Saxena, Aseem and Pandya, Harit and Kumar, Gourav and Gaud, Ayush and Krishna, K. Madhava},
  title     = {Exploring Convolutional Networks for End-to-End Visual Servoing},
  booktitle = {IEEE International Conference on Robotics and Automation (ICRA)},
  pages     = {3817--3823},
  year      = {2017}
}

@inproceedings{bateux2018training,
  author    = {Bateux, Quentin and Marchand, Eric and Leitner, J{\"u}rgen and Chaumette, Fran{\c{c}}ois and Corke, Peter},
  title     = {Training Deep Neural Networks for Visual Servoing},
  booktitle = {IEEE International Conference on Robotics and Automation (ICRA)},
  pages     = {3307--3314},
  year      = {2018}
}

@inproceedings{yu2019siamese,
  author    = {Yu, Cunjun and Cai, Zhongang and Pham, Hung and Pham, Quang-Cuong},
  title     = {Siamese Convolutional Neural Network for Sub-Millimeter-Accurate Camera Pose Estimation and Visual Servoing},
  booktitle = {IEEE/RSJ International Conference on Intelligent Robots and Systems (IROS)},
  pages     = {935--941},
  year      = {2019}
}

@inproceedings{harish2020dfvs,
  author    = {Harish, Y. V. S. and Pandya, Harit and Gaud, Ayush and Terupally, Shreya and Shankar, Sai and Krishna, K. Madhava},
  title     = {{DFVS}: Deep Flow Guided Scene Agnostic Image Based Visual Servoing},
  booktitle = {IEEE International Conference on Robotics and Automation (ICRA)},
  pages     = {9000--9006},
  year      = {2020}
}

@inproceedings{katara2021deepmpcvs,
  author    = {Katara, Pushkal and Harish, Y. V. S. and Pandya, Harit and Gupta, Abhinav and Sanchawala, AadilMehdi and Kumar, Gourav and Bhowmick, Brojeshwar and Krishna, K. Madhava},
  title     = {{DeepMPCVS}: Deep Model Predictive Control for Visual Servoing},
  booktitle = {Conference on Robot Learning (CoRL)},
  pages     = {2006--2015},
  year      = {2021}
}

@inproceedings{puang2020kovis,
  author    = {Puang, En Yen and Tee, Keng Peng and Jing, Wei},
  title     = {{KOVIS}: Keypoint-Based Visual Servoing with Zero-Shot Sim-to-Real Transfer for Robotics Manipulation},
  booktitle = {IEEE/RSJ International Conference on Intelligent Robots and Systems (IROS)},
  pages     = {7527--7533},
  year      = {2020}
}

@article{adrian2022dfbvs,
  author  = {Adrian, Nicholas and Do, Van-Thach and Pham, Quang-Cuong},
  title   = {{DFBVS}: Deep Feature-Based Visual Servo},
  journal = {arXiv preprint arXiv:2201.08046},
  year    = {2022}
}

@inproceedings{chen2024cns,
  author    = {Chen, Anzhe and Yu, Hongxiang and Wang, Yue and Xiong, Rong},
  title     = {{CNS}: Correspondence Encoded Neural Image Servo Policy},
  booktitle = {IEEE International Conference on Robotics and Automation (ICRA)},
  pages     = {17410--17416},
  year      = {2024}
}

@article{chen2025cnsv2,
  author  = {Chen, Anzhe and Yu, Hongxiang and Li, Shuxin and Chen, Yuxi and Zhou, Zhongxiang and Sun, Wentao and Xiong, Rong and Wang, Yue},
  title   = {{CNSv2}: Probabilistic Correspondence Encoded Neural Image Servo},
  journal = {arXiv preprint arXiv:2503.00132},
  year    = {2025}
}

@article{vitvs2025,
  author  = {Scherl, Alessandro and Thalhammer, Stefan and Neuberger, Bernhard and W{\"o}ber, Wilfried and Garc{\'i}a-Rodr{\'i}guez, Jos{\'e}},
  title   = {{ViT-VS}: On the Applicability of Pretrained Vision Transformer Features for Generalizable Visual Servoing},
  journal = {arXiv preprint arXiv:2503.04545},
  year    = {2025}
}

@misc{artvs2026,
  title         = {ART-VS: Adaptive Resolution Tiling for Vision Transformer Visual Servoing},
  author        = {Alessandro Scherl and Bernhard Neuberger and Simon Schwaiger and David Mulero-Pérez and Lucas Muster and Jose Garcia-Rodriguez},
  year          = {2026},
  eprint        = {2606.19089},
  archiveprefix = {arXiv},
  primaryclass  = {cs.RO},
  url           = {https://arxiv.org/abs/2606.19089}
}

@inproceedings{dust3r2024,
  author    = {Wang, Shuzhe and Leroy, Vincent and Cabon, Yohann and Chidlovskii, Boris and Revaud, J{\'e}r{\^o}me},
  title     = {{DUSt3R}: Geometric {3D} Vision Made Easy},
  booktitle = {IEEE/CVF Conference on Computer Vision and Pattern Recognition (CVPR)},
  pages     = {20697--20709},
  year      = {2024}
}

@inproceedings{mast3r2024,
  author    = {Leroy, Vincent and Cabon, Yohann and Revaud, J{\'e}r{\^o}me},
  title     = {Grounding Image Matching in {3D} with {MASt3R}},
  booktitle = {European Conference on Computer Vision (ECCV)},
  pages     = {71--91},
  year      = {2024}
}

@inproceedings{spann3r2024,
  author    = {Wang, Hengyi and Agapito, Lourdes},
  title     = {{3D} Reconstruction with Spatial Memory},
  booktitle = {International Conference on 3D Vision (3DV)},
  pages     = {78--89},
  year      = {2025}
}

@inproceedings{cut3r2025,
  author    = {Wang, Qianqian and Zhang, Yifei and Holynski, Aleksander and Efros, Alexei A. and Kanazawa, Angjoo},
  title     = {Continuous {3D} Perception Model with Persistent State},
  booktitle = {IEEE/CVF Conference on Computer Vision and Pattern Recognition (CVPR)},
  pages     = {10510--10522},
  year      = {2025}
}

@inproceedings{vggt2025,
  author    = {Wang, Jianyuan and Chen, Minghao and Karaev, Nikita and Vedaldi, Andrea and Rupprecht, Christian and Novotny, David},
  title     = {{VGGT}: Visual Geometry Grounded Transformer},
  booktitle = {IEEE/CVF Conference on Computer Vision and Pattern Recognition (CVPR)},
  pages     = {5294--5306},
  year      = {2025}
}

@inproceedings{pi32025,
  author    = {Wang, Yifan and Zhou, Jianjun and Zhu, Haoyi and Chang, Wenzheng and Zhou, Yang and Li, Zizun and Chen, Junyi and Pang, Jiangmiao and Shen, Chunhua and He, Tong},
  title     = {$\pi^3$: Permutation-Equivariant Visual Geometry Learning},
  booktitle = {International Conference on Learning Representations (ICLR)},
  year      = {2026}
}

@inproceedings{mapanything2025,
  author    = {Keetha, Nikhil and M{\"u}ller, Norman and Sch{\"o}nberger, Johannes and Porzi, Lorenzo and Zhang, Yuchen and Fischer, Tobias and Knapitsch, Arno and Zauss, Duncan and Weber, Ethan and Antunes, Nelson and Luiten, Jonathon and Lopez-Antequera, Manuel and Rota Bul{\`o}, Samuel and Richardt, Christian and Ramanan, Deva and Scherer, Sebastian and Kontschieder, Peter},
  title     = {{MapAnything}: Universal Feed-Forward Metric {3D} Reconstruction},
  booktitle = {International Conference on 3D Vision (3DV)},
  year      = {2026}
}

@article{vggtdp2025,
  author  = {Ge, Shijia and Liu, Yijun and Zhang, Yinxin and Xie, Shuzhao and Zhang, Weixiang and Zhou, Mingcai and Wang, Zhi},
  title   = {{VGGT-DP}: Generalizable Robot Control via Vision Foundation Models},
  journal = {arXiv preprint arXiv:2509.18778},
  year    = {2026}
}

@article{gp32025,
  author  = {Qian, Quanhao and Zhao, Guoyang and Zhang, Gongjie and Wang, Jiuniu and Xu, Ran and Gao, Junlong and Zhao, Deli},
  title   = {{GP3}: A {3D} Geometry-Aware Policy with Multi-View Images for Robotic Manipulation},
  journal = {arXiv preprint arXiv:2509.15733},
  year    = {2025}
}

@article{omnivggt2025,
  author  = {Peng, Haosong and Li, Hao and Dai, Yalun and Lan, Yushi and Luo, Yihang and Qi, Tianyu and Zhang, Zhengshen and Zhan, Yufeng and Zhang, Junfei and Xu, Wenchao and Liu, Ziwei},
  title   = {{OmniVGGT}: Omni-Modality Driven Visual Geometry Grounded Transformer},
  journal = {arXiv preprint arXiv:2511.10560},
  year    = {2025}
}

@inproceedings{mast3rslam2025,
  author    = {Murai, Riku and Dexheimer, Eric and Davison, Andrew J.},
  title     = {{MASt3R-SLAM}: Real-Time Dense {SLAM} with {3D} Reconstruction Priors},
  booktitle = {IEEE/CVF Conference on Computer Vision and Pattern Recognition (CVPR)},
  pages     = {16695--16705},
  year      = {2025}
}

@article{relocvggt2025,
  author  = {Deng, Tianchen and Wu, Wenhua and Wu, Kunzhen and Wang, Guangming and Zhu, Siting and Yuan, Shenghai and Chen, Xun and Shen, Guole and Liu, Zhe and Wang, Hesheng},
  title   = {{Reloc-VGGT}: Visual Re-Localization with Geometry Grounded Transformer},
  journal = {arXiv preprint arXiv:2512.21883},
  year    = {2025}
}

@article{wang2026vggtomega,
  author  = {Wang, Jianyuan and Chen, Minghao and Zhang, Shangzhan and Karaev, Nikita and Sch{\"o}nberger, Johannes and Labatut, Patrick and Bojanowski, Piotr and Novotny, David and Vedaldi, Andrea and Rupprecht, Christian},
  title   = {{{VGGT}-$\Omega$}},
  journal = {arXiv preprint arXiv:2605.15195},
  year    = {2026}
}

@article{umeyama1991least,
  author  = {Umeyama, Shinji},
  title   = {Least-Squares Estimation of Transformation Parameters Between Two Point Patterns},
  journal = {IEEE Transactions on Pattern Analysis and Machine Intelligence},
  volume  = {13},
  number  = {4},
  pages   = {376--380},
  year    = {1991},
  doi     = {10.1109/34.88573}
}

@inproceedings{kendall17_pose_regression,
  author    = {Kendall, Alex and Cipolla, Roberto},
  booktitle = {2017 IEEE Conference on Computer Vision and Pattern Recognition (CVPR)},
  title     = {Geometric Loss Functions for Camera Pose Regression with Deep Learning},
  year      = {2017},
  volume    = {},
  number    = {},
  pages     = {6555-6564},
  doi       = {10.1109/CVPR.2017.694}
}

@article{dinov3,
  title={Dinov3},
  author={Oriane Siméoni and Huy V. Vo and Maximilian Seitzer and Federico Baldassarre and Maxime Oquab and Cijo Jose and Vasil Khalidov and Marc Szafraniec and Seungeun Yi and Michaël Ramamonjisoa and Francisco Massa and Daniel Haziza and Luca Wehrstedt and Jianyuan Wang and Timothée Darcet and Théo Moutakanni and Leonel Sentana and Claire Roberts and Andrea Vedaldi and Jamie Tolan and John Brandt and Camille Couprie and Julien Mairal and Hervé Jégou and Patrick Labatut and Piotr Bojanowski},
  journal={arXiv preprint arXiv:2508.10104},
  year={2025}
}

\end{document}